\documentclass[english]{cccconf}
\usepackage[comma,numbers,square,sort&compress]{natbib}
\usepackage{epstopdf}
\usepackage{graphics} % for pdf, bitmapped graphics files
\usepackage{booktabs}
\usepackage{epsfig} % for postscript graphics files
\usepackage{graphicx}
\usepackage{multirow}
\usepackage{subfigure}
\usepackage{float}
\usepackage{amsmath} % assumes amsmath package installed
\usepackage{amssymb}  % assumes amsmath package installed
\usepackage{tabularx}
\usepackage{threeparttable} % add footnote for table
\usepackage[colorlinks, linkcolor=red, citecolor=blue]{hyperref}

\begin{document}

\title{Learning adaptive manipulation of objects with revolute joint: A case study on varied cabinet doors opening}

\author{Hongxiang Yu\aref{amss},  Dashun Guo\aref{amss}, Zhongxiang Zhou\aref{amss}, Yue Wang\aref{amss} and Rong Xiong\aref{amss}}

% Note: the first argument in the \affiliation command is optional.
% It defines a label for the affiliation which can be used in the \aref
% command. If there is only one affiliation for all authors, then the
% optional argument in the \affiliation command should be suppressed,
% and the \aref command should also be removed after each author in
% \author command, in this case the affiliation will not be numbered.

% Çë×¢Òâ£º\affiliationÃüÁîµÄµÚÒ»¸ö²ÎÊýÊÇ¿ÉÑ¡µÄ£¬Ëü¶¨ÒåÁËÓÃÓÚ\arefÃüÁîµÄ±êÇ©¡£
% Èç¹ûËùÓÐ×÷ÕßÖ»ÓÐÒ»¸öµ¥Î»£¬Çë²»ÒªÊ¹ÓÃ\affiliationÃüÁîµÄ¿ÉÑ¡²ÎÊý£¬Í¬Ê±ÔÚÉÏÃæ
% µÄ\authorÃüÁîÖÐµÄÃ¿Î»×÷ÕßÐÕÃûºóÃæÒ²²»ÄÜÊ¹ÓÃ\arefÃüÁî£¬Ê¾ÀýÈçÏÂ
% \author{First Author, Second Author, Third Author}
% \affiliation{Chinese Academy of Sciences, Beijing 100190, P.~R.~China\email{ccc@amss.ac.cn}}
% ´ËÊ±µ¥Î»Ç°²»»áÓÐÊý×Ö±àºÅ£¬×÷ÕßÐÕÃûºóÃæÒ²Ã»ÓÐ±àºÅ

\affiliation[amss]{the State Key Laboratory of Industrial Control and Technology, and the Institute of Cyber-Systems and Control, Zhejiang University, Hangzhou 310058, China.
        \email{hongxiangyu@zju.edu.cn}}
% \affiliation[hit]{Harbin Institute of Technology, Harbin 150001, P.~R.~China
%         \email{xxx@hit.edu.cn}}

\maketitle

\begin{abstract}
This paper introduces a learning-based framework for robot adaptive manipulating the object with a revolute joint in unstructured environments. We concentrate our discussion on various cabinet door opening tasks. To improve the performance of Deep Reinforcement Learning in this scene, we analytically provide an efficient sampling manner utilizing the constraints of the objects. To open various kinds of doors, we add encoded environment parameters that define the various environments to the input of out policy. To transfer the policy into the real world, we train an adaptation module in simulation and fine-tune the adaptation module to cut down the impact of the policy-unaware environment parameters. We design a series of experiments to validate the efficacy of our framework. Additionally, we testify to the model's performance in the real world compared to the traditional door opening method.
\end{abstract}


% Please remove or comment out the following line if the footnote is not necessary
% \footnotetext{This work is supported by National Natural Science
% Foundation (NNSF) of China under Grant 00000000.}

\footnotetext{This work was supported in part by the National Key R$\&$D Program of China under Grant 2021ZD0114500 and the National Nature Science Foundation of China under Grant 62173293.}

\section{Introduction}

Robots working in human environments often encounter a wide range of objects with revolute joints, such as cabinet doors, glasses, and laptops, which contain multiple kinematically linked functional bodies. Developing robots capable of interacting with those instances safely faces two main challenges. (1) Manipulation of objects with revolute joint relies heavily on the accurate kinematic model or pose estimation so that imprecise action is unsafe for the revolute joints, especially when the robot arm and the object fully contact. (2) Robots working in the unstructured and dynamic environment need the capability to operate with various and unknown objects, which requires the policy to be general to the new objects without task-specific training data.

Existing research is mainly divided into two routes. Some works focus on the speed control. Traditional methods with speed control \cite{nemec2017door} typically abstract 6D axis poses from visual observations and then plan on top of the inferred poses analytically. These methods rely heavily on accurate pose estimation and struggle to generalize to new objects without extra data. Recently, other researchers focus on learning policies from demonstration by leveraging experience from human \cite{englert2018learning}\cite{eteke2020reward}, which achieves significant progress. However, as the unknown object has different dynamic models, data distribution shift from the demonstrated objects to the inference objects still remains, which results in poor performance on the unknown objects. 

Works in another line focus on close-loop control with force-torque data as feedback. \cite{karayiannidis2016adaptive} leverages the wrist force-torque data to estimate the kinematic model, which can generalize to the unknown objects without extra data. However, the generalization ability is limited as the dynamic parameters of the objects (e.g., friction and mass) are out of consideration. Moreover, the policy would be unstable with improper initialization.

\begin{figure}[t]
\centering
\includegraphics[width=\linewidth]{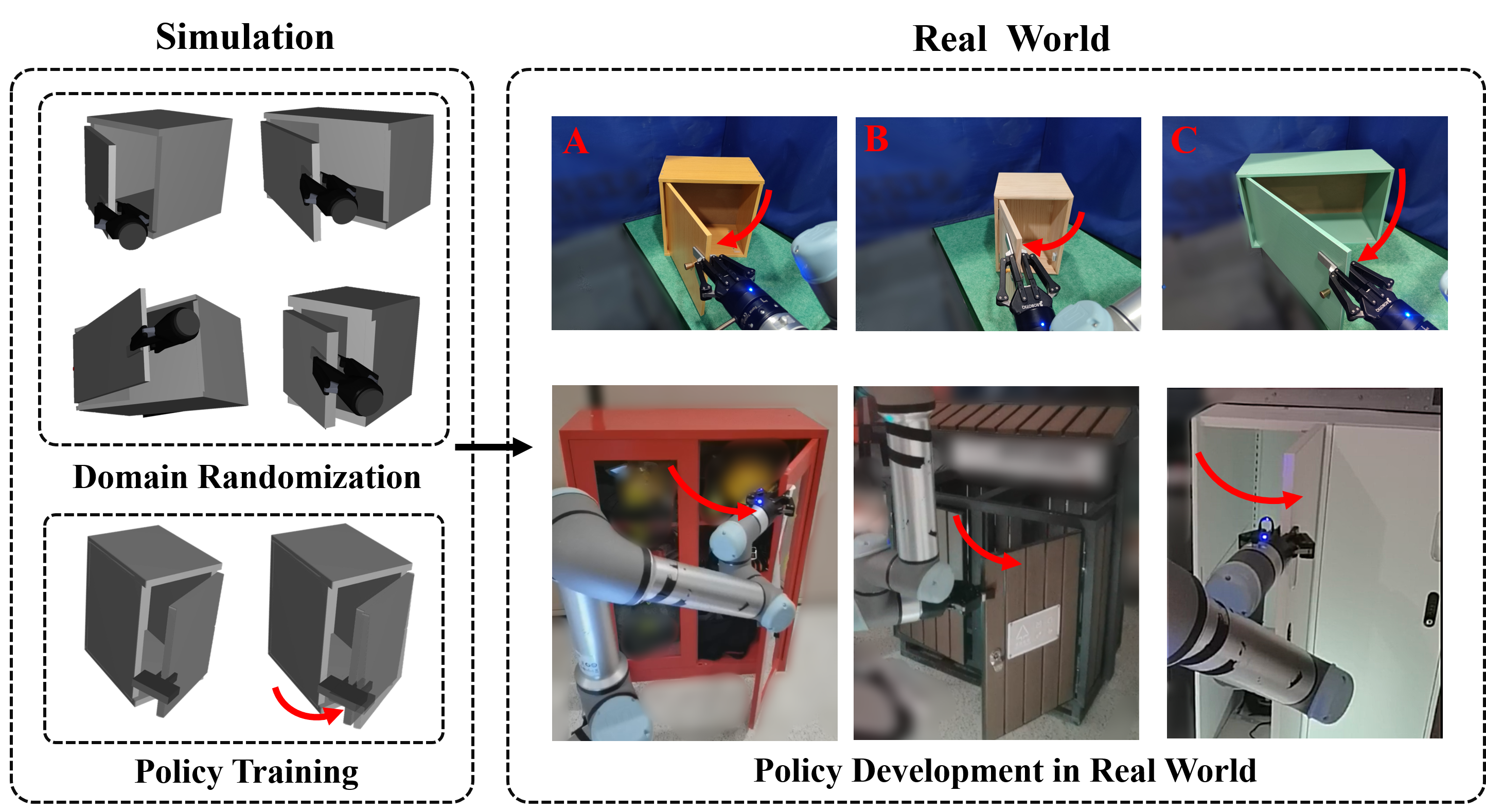}
\caption{We provide a learning-based framework for revolute joint objects manipulation. With domain randomization and an adaptation module, we are able to open various kinds of cabinet doors in the real world with the policy trained in simulation. We then implement our real-world experiments on the cabinet doors A, B, C and compare our method with the traditional door opening method.}
\label{teaser}
\vspace{-0.5cm}
\end{figure}

% \begin{figure*}[ht]
%     \centering
%     \includegraphics[width=0.8\linewidth]{fig/frame_0.png}
%     \caption{Overview of our method. Dimension reduction of the searching space is the foundation of the policy training. We train the encoder to modulate the environment parameters. Then we train the policy based on the domain randomization. The encoder and the policy make up our base policy (BP). Then, we propose an adaptation module to estimate the parameters of the unknown cabinets. We train the adaptation module with the supervision of the encoder. The adaptation module and the policy make up our adaptive policy (AP). With the supervision of the base policy, we fine-tune the adaptation module and obtain the fine-tuned adaptive policy (FAP) for real world door opening tasks.}
%     \label{fig:overview}
% \vspace{-0.4cm}
% \end{figure*}

\begin{figure*}[t]
\centering

    \subfigure[Coordinates Definition]{
    \includegraphics[width=0.23\linewidth]{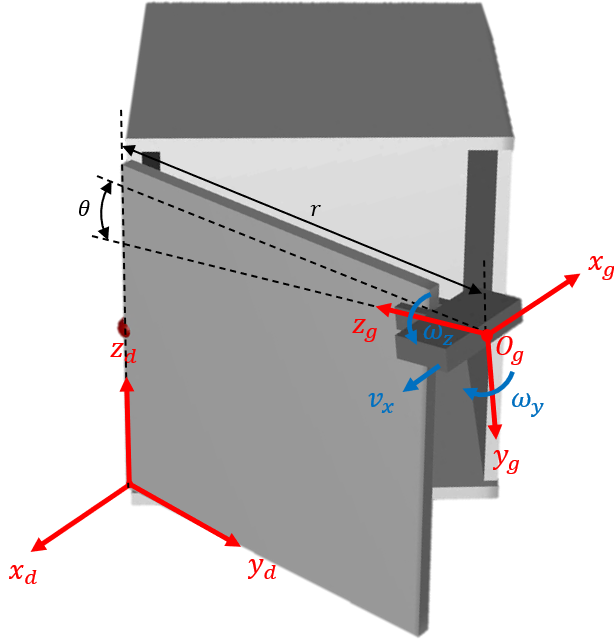}
    \label{space}
    }
    \subfigure[Training Pipeline]{
    \includegraphics[width=0.73\linewidth]{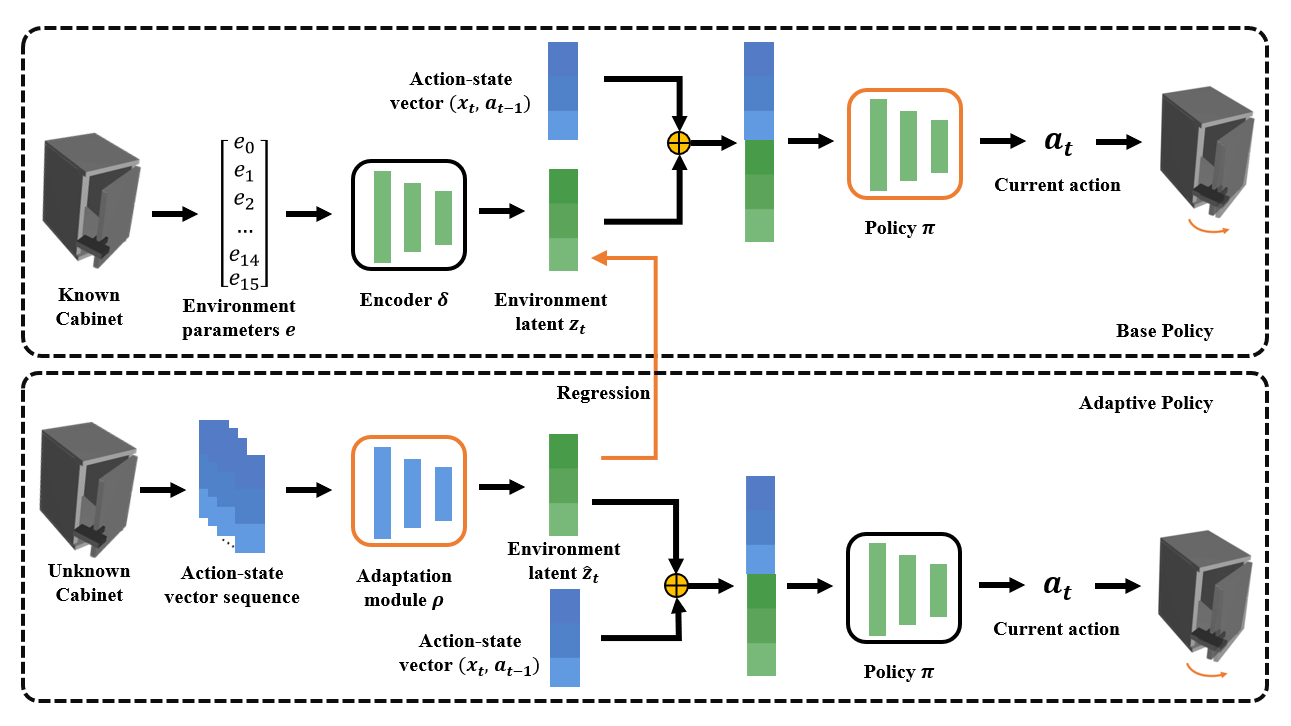}
    \label{frame1}
    }

\caption{The left part shows the coordinates definition of the dimension reduction. The right part shows our training pipeline. The first line is the train process of our policy. Given the environment latent vector $\boldsymbol{z_t}$, the current state $\boldsymbol{x_t}$ and the current state $\boldsymbol{s_t}$ of the environment, the policy is able to open the known cabinet door. The encoder and the policy make up the base policy (BP). The second line is the adaptation module to estimate the environment latent vector $\boldsymbol{\hat{z}_t}$ with the action-state sequence. We train the adaptation module with the supervision of the environment latent vector $\boldsymbol{z_t}$. The adaptation module and the policy make up the adaptive policy (AP). The orange modules in the figure are trainable while the black ones are fix.}
\vspace{-0.5cm}
\end{figure*}

In this work, we use the Deep Reinforcement Learning (DRL) to learn a general policy to manipulate unseen objects by putting the force-torque data into the control loop. Compared with the methods based on speed control, we provide the system with force-torque feedback to eliminate the estimation error in real-time. On the other hand, compared with the methods based on force-position control, we improve the performance of the generalization on the unseen objects by considering the dynamic parameters of the objects. 

%We choose the force-torque data rather than the image data as the input of the policy, which develops a feedback control to avoid the inaccuracy of the initial estimation. 
%to replace complex dynamical modeling and policy hand-engineering. DRL is widely applied in manipulation tasks but often meets the challenge of the high sample complexity \cite{liu2021deep}. 
% The designed space guides the agent to explore an optimal solution rather than wildly scratch in the environment. 

%Training policy in the real world is practically infeasible, for the effector randomly interacting would cause damage to either the effector or the objects, so we train our base policy in the simulation.
%Then we create an environment to simulate various cabinet doors based on this domain and train the agent to interact with them. 
%We train the adaptation module in the simulation with the supervision of the encoded environment parameters. 
Generally, this study extends an adaptive manipulation of the object with a revolute joint and provides a case study on cabinet door opening.
We analytically reduce the dimension of the searching space to improve the sample efficiency of DRL. We also define a domain that contains the environment parameters of cabinet doors subject to daily life. After that, we 
generate a policy with encoded environment parameters as part of the input, which allows the policy to adapt to the given environment. 
However, we are not accessible to the accurate environment parameters in the real world, so we propose an adaptation module to estimate the parameters. 
Furthermore, we fine-tune the adaptation module to cut down the impact of the environment parameters which are policy-unaware. With the help of the adaptation module, the policy can be directly developed in the real world. A series of simulations and real-world experiments are conducted to validate our framework's efficacy and performance.

Our contributions are as follows:
\begin{itemize}
    \item We analytically utilize the constraints of the cabinet door to dilute the sample complexity of DRL.
    \item We decorate the input of our policy with encoded parameters that define the various environments and train an adaptation module for the real world deployment.
    \item We design a series of simulations and real-world experiments to validate our framework's efficacy, and we compare the performance of our policy with the traditional door opening method.
\end{itemize}

\section{RELATED WORK}

\subsection{Door Opening}

Our daily environments are commonly populated with multifarious doors. Lots of previous works related to door opening physically develop the required manipulation policy by the estimation of the door's detailed geometrical model \cite{nemec2017door}\cite{hausman2015active}\cite{liu2022towards}.    
ScrewNet \cite{jain2021screwnet} is a novel approach that estimates an object’s articulation model directly from depth images without requiring a priori knowledge of the articulation model category. \cite{abbatematteo2019learning} presents a
framework for estimating the kinematic model and configuration of articulated objects and interacting with the objects. Furthermore, \cite{xu2022umpnet} proposes a policy that generalizes to unseen objects or categories and then  applies the estimation results to the real-world door opening. However, these methods are not adaptive for the lack of real-time capabilities for handling dynamic changes.

Instead of obtaining the kinematical model, \cite{arduengo2021robust} combines a versatile Bayesian framework with a Task Space Region motion planner. They achieve an efficient door operation but rely on the collection of datasets. \cite{karayiannidis2016adaptive} proposes a methodology for simultaneous compliant interaction and estimation of constraints imposed by the joint, while the system they design would be unstable with improper initial values.

\subsection{Deep Reinforcement Learning for Manipulation}

Plenty of research in recent years proves that deep reinforcement learning has provided a way to enable robots to flexibly interact with varied objects in a broad range of tasks and unstructured  environments\cite{gu2017deep}\cite{popov2017data}. Compared with traditional control, DRL changes the pattern from hand-coding to data driving, while the apparent limitation is that collecting data from scratch is time-consuming. Some researchers develop the learning through demonstrations in imitation learning to achieve sample efficient DRL \cite{tai2016survey}. They put the manual data into a replay buffer to replace the policy initialization. However, they lack the capability of transferring to different task scenes and are constrained in trajectory-based policy representation. Other works allow large-scale data collection using several robots to gather experience in parallel \cite{levine2018learning}\cite{yahya2017collective}. However, their challenges are the accuracy for real applications due to the sensitivity to data and hardware.

\section{METHODS}

% The framework of our method is shown as Fig. \ref{fig:overview}. 
In Section \ref{subseactionsearching}, we analyze the dimension reduction of the searching space, which is the foundation of the policy training. In Section \ref{section:encodertraining} and \ref{section:policytraining}, we describe the environment parameters and the training process of the encoder. Then we train the policy based on the domain randomization. The encoder and the policy make up our base policy (BP). In Section \ref{section:adaptation}, we propose an adaptation module to estimate the parameters of the unknown cabinets. We train the adaptation module with the supervision of the encoder. The adaptation module and the policy make up our adaptive policy (AP). Finally, in section \ref{subsectionfine}, we discuss the relationship between the environment parameters and the policy. With the supervision of the base policy, we fine-tune the adaptation module and obtain the fine-tuned adaptive policy (FAP). We develop the fine-tuned adaptive policy in the real world directly for the door opening tasks.

% In this section, we demonstrate the components of our framework respectively. In Section \ref{subseactionsearching}, we analyze the dimension reduction of the searching space, and we convert the dimension from six to two, which improves the sample efficiency of DRL. In Section \ref{section:adaptivepolicy},
% we provide the details of our encoder, base policy, and adaptation module. We carry out all of the training processes in the simulation. In Section \ref{subsectionfine}, we discuss the relationship between the environment parameters and the policy. We fine-tune the adaptation module to cut down the impact of the policy-unaware environment parameters. Finally, we directly develop our base policy with the support of the fine-tuned adaptation module.

\subsection{Dimension Reduction of Searching Space}\label{subseactionsearching} 
% Rather than free-joint objects, objects with revolute joints have kinematical structures that cannot be violated, which means the effector randomly interacting would cause damage to either the effector or the objects. Therefore, searching for reinforcement learning in the whole action space is time-consuming and nonsense. Firstly, we contract the searching space by physically modeling the manipulation task.

According to Fig. \ref{space}, we make the following assumptions: (1) The effector is a robotic gripper and has grasped at the cabinet door. The YZ-plane of the gripper is parallel to the YZ-plane of the cabinet door and between z-axis of the gripper $\{z_g\}$ and y-axis of the cabinet door $\{y_d\}$ there is an angle $\theta$. (2) The distance from the gripper’s end center $O_g$ to the z-axis of the cabinet door $\{z_d\}$ is $r$. (3) In the ideal door opening process, the cabinet door rotates uniformly with the angular velocity 
\begin{equation}
    \boldsymbol{\omega}_d = \begin{bmatrix}
  0& 0 & \omega
\end{bmatrix}^T
\label{omega_d}
\end{equation}
while the gripper is stationary with respect to the cabinet door during the process.
We set the desired gripper’s velocity $\widetilde{\mathbf{v}}$ with respect to the gripper frame $\{g\}$ as 
\begin{equation}
    \mathbf{\widetilde{v}} =\omega \begin{bmatrix}
    r & 0 & 0 & 0 & -cos\theta &sin\theta
    \end{bmatrix}^T
\label{v_drtheta}
\end{equation}

Based on Eq.(\ref{v_drtheta}), we convert the searching space of DRL from $\mathbf{\widetilde{v}} \in \mathbb{R}^{6\times1}$ to $\begin{bmatrix}
r & \theta
\end{bmatrix}^T \in \mathbb{R}^{2\times1}$, which speeds up the model’s training process and raises up the model’s performance. For more details, please refer to the supplementary material.

\begin{figure}[t]
\centering
\includegraphics[width=0.9\linewidth]{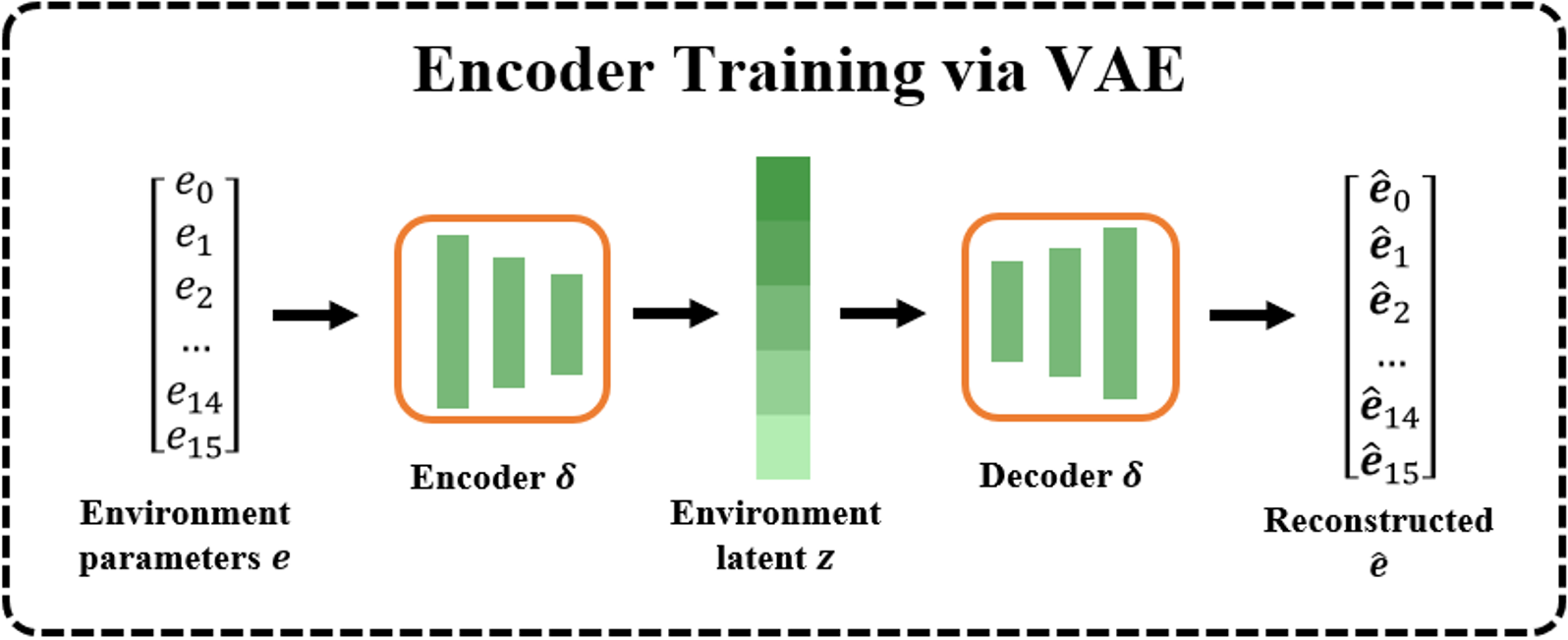}
\caption{A variational autoencoder to encode the environment parameters vector $\boldsymbol{e}$ into a low-dimension environment latent vector $\boldsymbol{z}$. The encoder $\delta$ of the VAE is a 3-layer multi-layer perception (MLP) (16,16 hidden layer sizes) and encodes $\boldsymbol{e}\in \mathbb{R}^{1\times16}$ into $\boldsymbol{z}\in \mathbb{R}^{1\times8}$. }
\label{fig:vae}
\vspace{-0.4cm}
\end{figure}

\subsection{Enviromment Parameters and Encoder Training}\label{section:encodertraining}

To depict the various configurations of the cabinet doors, we specify the environment parameters $\boldsymbol{e}$ subject to daily life, with their value range shown in the supplementary material. We create an environment and then train our policy in the environment based on the domain randomization. The training settings guarantee the generalization of our policy.

We pretrain a variational autoencoder (VAE) \cite{kingma2013auto} to encode the $\boldsymbol{e}$ into a low-dimension environment latent vector $\boldsymbol{z}$ (Fig. \ref{fig:vae}). The encoder $\delta$ of the VAE is a 3-layer multi-layer perception (MLP) (16,16 hidden layer sizes) and encodes $\boldsymbol{e}\in \mathbb{R}^{1\times16}$ into $\boldsymbol{z}\in \mathbb{R}^{1\times8}$.

As Eq. (\ref{equ:vaeloss})-(\ref{equ:vaelreg}) show, the loss function $l_e$ of VAE consists of two parts: the reconstruction loss $l^{rec}_e$ and the regular loss $l^{reg}_e$. The reconstruction loss $l^{rec}_e$ leads the estimation $\boldsymbol{\hat{e}}$ to converge to the input $\boldsymbol{e}$ and the regular loss $l^{reg}_e$ empowers the latent vector $\boldsymbol{z}$ conforming to an independent normal distribution.
\begin{equation}
    l_e = l^{rec}_e + l^{reg}_e
\label{equ:vaeloss}
\end{equation}
\begin{equation}
    l^{rec}_e = -||\boldsymbol{e}-\boldsymbol{\hat{e}}||^2
\label{equ:vaelrec}
\end{equation}
\begin{equation}
    l^{reg}_e =  -\frac{1}{2}\sum_{i=0}^{n} (\mu_i^2+\sigma_i^2-log\sigma_i^2-1)
\label{equ:vaelreg}
\end{equation}
where the $\boldsymbol{\mu}$, $\boldsymbol{\sigma}$ are the mean and standard deviation of $\boldsymbol{z}$.
A reparameterization trick is applied to sample the $\boldsymbol{z}$ as Eq.(\ref{equ:samplez}) describes, which makes the sampling process differentiable.
\begin{equation}
    \boldsymbol{\mu}, \boldsymbol{\sigma}= \delta(\boldsymbol{e})
\label{equ:encodemusigma}
\end{equation}
\begin{equation}
    \boldsymbol{z} = \boldsymbol{\mu} + \boldsymbol{\epsilon} \odot \boldsymbol{\sigma}
\label{equ:samplez}
\end{equation}
where $\boldsymbol{\epsilon} \sim \mathcal N(0, I_d)$.

Consequently, we condense the environment parameter $\boldsymbol{e}$ into the smaller environmental latent vector $\boldsymbol{z}$, which dilutes the training cost of the base policy. Furthermore, the input conforming to a normal distribution speeds up the training process by the more significant gradient.

\subsection{Policy Training}\label{section:policytraining}
The door opening policy $\pi$ is trained by proximal policy optimization (PPO) \cite{schulman2017proximal}. The input of policy is the current state $\boldsymbol{x_t}\in \mathbb{R}^{1\times22}$, comprised of the environment latent vector  $\boldsymbol{z_t}\in \mathbb{R}^{1\times8}$, the state $\boldsymbol{s_t}\in \mathbb{R}^{1\times12}$ of the environment, and the previous action $\boldsymbol{a_{t-1}}\in \mathbb{R}^{1\times2}$, the output is the target action $\boldsymbol{a_{t}}$. The current state consists of the force torque sensor’s data and the gripper’s velocity with respect to the world frame $\{w\}$. The action $\boldsymbol{a}$ consists of the estimation $\hat{r}$ and $\hat{\theta}$ as we discussed in Section \ref{subseactionsearching}:
\begin{equation}
    \boldsymbol{a} = \pi(\boldsymbol{x})
\label{basepolicy}
\end{equation}

We encourage the agent to generate the correct $r$ and $\theta$ through the rewards function. In the simulation, the truth $r$ and $\theta$ are available so that the first part of the reward function is a supervising impetus with the mean squared error (MSE) between $[r,\theta]$ and $[\hat{r}, \hat{\theta}]$. It guides the agent to track the ideal trail of the door opening.

Unfortunately, the supervising part is insufficient, for the agent feels confused facing a situation it has never met before. We propose another feedback impetus proportional to the force-torque sensor’s data $\begin{bmatrix}
F_x & F_y & F_z & \tau_x & \tau_y & \tau_z
\end{bmatrix}$ with respect to the frame $\{g\}$. Assuming in the ideal case that gripper drives the cabinet door to rotate at a constant angular velocity, it receives the force $\boldsymbol{F_{dg}} = \begin{bmatrix}
-F & 0 & 0
\end{bmatrix}$ and torque $\boldsymbol{\tau_{dg}} = \begin{bmatrix}
0 & 0 & \tau
\end{bmatrix}$ with respect to the frame $\{d\}$ $(F, \tau>0)$. In the frame $\{g\}$, the force $\boldsymbol{F_g}$ and torque $\boldsymbol{\tau_g}$ are
\begin{equation}
\boldsymbol{F_g} = \begin{bmatrix}
 F & 0 &0
\end{bmatrix}
\label{f_g}
\end{equation}
\begin{equation}
    \boldsymbol{\tau_g}=\begin{bmatrix}
0& -\tau cos\theta & \tau sin\theta
\end{bmatrix}
\label{tau_g}
\end{equation}

Eq. (\ref{f_g}) and (\ref{tau_g}) indicate that the gripper consecutively suffers a force in axis $\{x_g\}$ and a torque with a constant angle $\theta$ in axis $\{y_g\}$ and axis $\{z_g\}$.
Consequently, the reward function at time $t$ is defined as
\begin{equation}
\begin{aligned}
      R(\boldsymbol{x_t}, \boldsymbol{a_{t-1}}) =  - & [k_1(||r-\hat{r}||^2+||\theta-\hat{\theta}||^2) +  \\ k_2|F_y| +  &  k_3|F_z| +   k_4|\tau_x| + k_5|\frac{\tau_z}{\tau_y}+tan\theta|]  
\end{aligned}
\label{reward_rt_new}
\end{equation}
where $k_{1:5}$ are hyper-parameters in the training process.

\subsection{Adaptation Module Training}\label{section:adaptation}
According to Eq. (\ref{equ:encodemusigma}) and (\ref{equ:samplez}), the policy acquires the $\boldsymbol{z_t}$ from the $\boldsymbol{e_t}$ as a part of the state $\boldsymbol{x_t}$. While there are convenient interfaces to access to the $\boldsymbol{e_t}$ in the simulation, it’s impracticable to obtain the accurate $\boldsymbol{e_t}$ in the real world. We propose an adaptation module $\rho$ to estimate the $\boldsymbol{\hat{z}_t}$. We set the module's input as the action and state sequence based on the assumption that the same effector's action gives the same feedback in the same environment. We choose to estimate $\boldsymbol{z_t}$ through the adaptation module $\rho$ rather than $\boldsymbol{e_t}$, because our goal is not to identify a specific system but to obtain a correct action. 

We train the adaptation module with on-policy data as shown in the bottom line of Fig. \ref{frame1}. In the simulation, we can obtain $\boldsymbol{e_t}$, and then obtain the ground truth value $\boldsymbol{z_t}$ through $\delta$. At the same time, we can record the action state sequence and then obtain the estimated $\boldsymbol{\hat{z_t}}$ through $\rho$. It inspires us that we can train the module with the supervision of $\boldsymbol{z_t}$ from the encoder.
According to Eq. (\ref{equ:vaeloss})-(\ref{equ:vaelreg}), the loss function $l_a$ of the adaptation module is similarly defined as
\begin{equation}
    l_a = l^{rec}_a + l^{reg}_a
\label{equ:adploss}
\end{equation}
\begin{equation}
    l^{rec}_a = -(||\boldsymbol{\mu}-\boldsymbol{\hat{\mu}}||^2+||\boldsymbol{\sigma}-\boldsymbol{\hat{\sigma}}||^2)
\label{equ:adplrec}
\end{equation}
\begin{equation}
    l^{reg}_a =  -\frac{1}{2}\sum_{i=0}^{n} (\hat{\mu}_i^2+\hat{\sigma}_i^2-log\hat{\sigma}_i^2-1)
\label{equ:adplreg}
\end{equation}
where $\boldsymbol{\hat{\mu}}$, $\boldsymbol{\hat{\sigma}}$ are the mean and standard deviation of $\hat{z}$.

And according to the Eq. (\ref{equ:encodemusigma}) and (\ref{equ:samplez}), the $\boldsymbol{\hat{z}}$ is sampled in the same way as
\begin{equation}
    \boldsymbol{\hat{\mu}_t}, \boldsymbol{\hat{\sigma}_t}= \rho(\boldsymbol{x_{t-n:t}})
\label{equ:adpmusigma}
\end{equation}
\begin{equation}
    \boldsymbol{\hat{z_t}} = \boldsymbol{\hat{\mu_t}} + \boldsymbol{{\epsilon}'} \odot \boldsymbol{\hat{\sigma_t}}
\label{equ:adpsamplez}
\end{equation}
where $\boldsymbol{{\epsilon}'} \sim \mathcal N(0, I_d)$.

%We choose MSE as the loss function in the training process.  

The input of the model $\boldsymbol{x_{t-n:t}} \in \mathbb{R}^{n\times26}$ includes the state $\boldsymbol{s_{t-n:t}}\in \mathbb{R}^{n\times12}$ of the environment and the past action $\boldsymbol{a_{t-n-1:t-1}}\in \mathbb{R}^{n\times2}$, the output is the estimated value of environment latent vector $\boldsymbol{{\hat{z}}_t}$. The first part of the adaptation module is a 2-layer MLP. And then, we add 3-layer 1D-CNN to $\rho$ to extract the temporal information from the action-state sequence. The input channel
number, output channel number, kernel size, and stride of each
layer are $[32, 32, 8, 4], [32, 32, 5, 1], [32, 32, 5, 1]$. Finally, there is a liner layer to flatten the output of $\rho$ projected to estimate $\boldsymbol{{\hat{z}}_t}$.

\subsection{Fine-tuning for Adaptation Module}\label{subsectionfine}

There is no solid one-to-one correspondence between state $\boldsymbol{x}$ action $\boldsymbol{a}$ and the environment parameters $\boldsymbol{e}$. In other words, some factors of the environment parameters make the environment variables, but they may not impact the output of the policy. 
Hence the adaptation module cannot estimate accurate $\boldsymbol{z}$ with the action-state sequence as its input. 

% Changes in any factor of the environment parameter $\boldsymbol{e}$ make the environment variables, but they may not impact the output of the policy. Conversely, the same $\boldsymbol{x}$ and $\boldsymbol{a}$ are probably correspond to different $\boldsymbol{e}$. That means the adaptation module cannot estimate accurate $\boldsymbol{z}$ with the action-state sequence as its input. 

In the section, we divide $\boldsymbol{z}$ into policy-aware $\boldsymbol{z^a}$ and policy-unaware $\boldsymbol{z^u}$ based on whether it affects the policy’s output. Accordingly, the factors $\boldsymbol{z^a}$ result in policy-aware $\boldsymbol{\hat{z}^a}$ after the module $\rho$ and the policy-unaware factors $\boldsymbol{z^u}$ lead to noise $\boldsymbol{\hat{z}^u}$ which worse the policy’s performance.

\begin{figure}[t]
\centering
\includegraphics[width=0.9\linewidth]{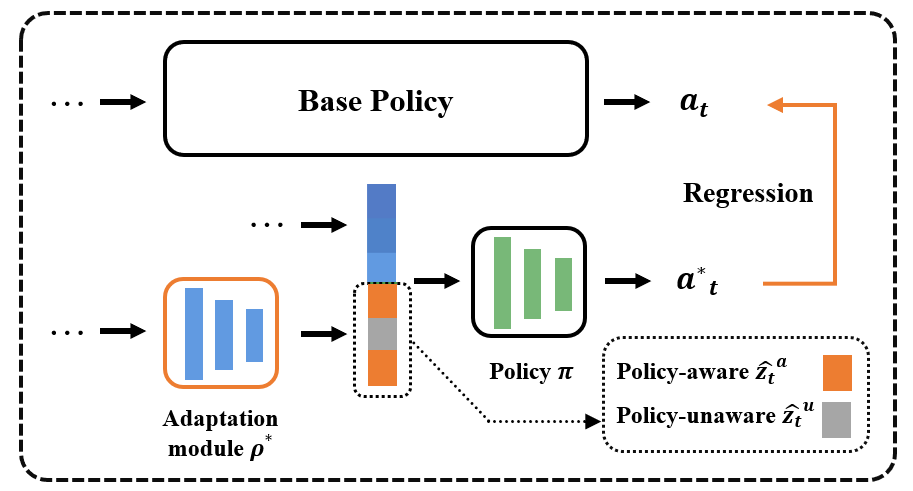}
\caption{Fine-tuning Process for the adaptation module. We divide the environment latent vector $\boldsymbol{z_t}$ into policy-aware $\boldsymbol{z^a}$ and policy-unaware $\boldsymbol{z^u}$, which are recolored by orange and grey in the figure. We obtain a new action $\boldsymbol{a^*}$ through the new adaptation module $\rho^*$. With the supervision of the action $\boldsymbol{a}_t$ related to $\boldsymbol{z_t}$, we fine-tune the adaptation module and cut down the impact on the policy of $\boldsymbol{z^u}$.}
\label{frame2}
\vspace{-0.7cm}
\end{figure}

Assuming that the new adaptation module $\rho^*$ produces the new estimation $\boldsymbol{\hat{z}^*}$, and according to the Eq. (\ref{basepolicy})(\ref{equ:adpmusigma})(\ref{equ:adpsamplez}) we get the related action $\boldsymbol{a^*}$ (Fig. \ref{frame2}). Our goal now is fine-tune the adaptation module though fitting $\boldsymbol{a^*}$ to $\boldsymbol{a}$ to eliminate the impact of noise on the policy. Therefore, the loss function $l_f$ of this stage is defined as 

\begin{equation}
    l_f = ||\boldsymbol{a}-\boldsymbol{a^*}||^2
\label{lossfunctionf}
\end{equation}
In the training process, we detach the gradient of the policy $\pi$, preventing the policy's weight updating during the gradient backpropagation.

% At this moment, we train a door opening policy and adapt it to a given cabinet door with environment parameters. To obtain the environment parameters of an unknown cabinet door, we propose an adaptation module to estimate the environment latent vector in real-time. We fine-tune the adaptation module to cut down the impact of the policy-unaware environment parameters. 
% we train all components of FAP in the simulation, but

Consequently, we directly develop the FAP in the real world with the fine-tuned adaptation module's support. Section \ref{section:results} provides a detailed discussion of  FAP's performance in the real world.

\section{EXPERIMENT SETUP}

\subsection{Simulation Setup}
We train the policy in the simulation environments containing a variety of cabinet doors powered by MuJoCo \cite{6386109}. Inspired by the dataset \cite{abbatematteo2019learning}, we generate cabinet doors with different properties by randomizing the parameters in XML model files. We simulate an individual Robotiq 2F-140 gripper without other robotic arm joints so that we directly set the velocity of the gripper rather than the inverse kinematic calculation. The gripper is initially attached to the cabinet door as stated in assumption (1). We set the max timestep of each episode to be 500, and the control frequency of the simulation is 100Hz.

\subsection{Real World Setup}

We use the UR5 robot for our real-world experiments, which functionally supports velocity control. We equip it with an Robotiq FT300 force-torque sensor for the requirement of the force-torque data. We choose the end effector is an Robotiq 2F-140 gripper, which allows us to grasp the cabinet doors. In order to meet assumption (1), in each experiment, we roughly attach the gripper to the cabinet door and then adjust it with a PI controller. We provide three cabinet doors with different properties to validate the generalization of our method. 

\subsection{Other Details}
All the training processes of the models and simulation experiments are carried out on a server with 16 Intel(R) Core(TM) i9-9900K 3.60GHz CPU and one NVIDIA GeForce RTX2080 SUPER GPU. Meanwhile, the real-world experiments are performed on a computer with 12 Intel(R) Core(TM) i7-8700 3.20GHz CPU and one NVIDIA GeForce GTX1060 GPU. The hyper-parameters of the reward function are set as $k_1=1, k_2=k_3=k_4=0.7, k_5=0.3$. The learning rates of VAE, PPO and the adaptation module are 1e-3, 3e-4, and 1e-4, respectively.

% table1
\begin{table}[t]
\centering
\caption{Single Door VS. Domain Randomization}
\label{singlevsdomain}
\setlength{\tabcolsep}{10pt}
\begin{threeparttable}

\begin{tabular}{lcccc}
\toprule[1pt]

                     & \multicolumn{4}{c}{\textbf{Error}(*$e^{-2}$)}                                                      \\
                     & \multicolumn{2}{c}{\textbf{r}(m)}                             & \multicolumn{2}{c}{$\boldsymbol{\theta}$(rad)}          \\

                     & \textbf{Mean}          & \textbf{90}\tnote{1} & \textbf{Mean}          & \textbf{90}\tnote{1}  \\
                     \midrule
\textbf{SD}\tnote{2}          & 15.8          & 31.1                              & 13.0          & 22.6                               \\
\textbf{DR}\tnote{3} & \textbf{4.56} & \textbf{12.7}                     & \textbf{2.17} & \textbf{4.41}                      \\
\bottomrule[1pt]
\end{tabular}

\begin{tablenotes}
\footnotesize
\item[1] 90 means the $90^{th}$ percentile of the data.
\item[2] Policy training on single door.
\item[3] Policy training by domain randomzation.
\end{tablenotes}
\end{threeparttable}
\vspace{-0.4cm} \end{table}

% table2
\begin{table}[t]
\centering
\caption{Policy With Encoder VS. Policy Without Encoder}
\label{encoderwithout}
\setlength{\tabcolsep}{10pt}
\begin{threeparttable}
\begin{tabular}{lcccc} 
\toprule[1pt]
& \multicolumn{4}{c}{\textbf{Error}(*$e^{-2}$)}                                                      \\
& \multicolumn{2}{c}{\textbf{r}(m)}                             & \multicolumn{2}{c}{$\boldsymbol{\theta}$(rad)}          \\
& \textbf{Mean}          & \textbf{90}\tnote{1} & \textbf{Mean}          & \textbf{90}\tnote{1}  \\
\midrule
\textbf{WoE}\tnote{2}          & 9.54         & 20.2                             & 9.90         & 20.0                               \\
\textbf{WE}\tnote{3}              & \textbf{4.56} & \textbf{12.7}                     & \textbf{2.17} & \textbf{4.41}                      \\
\bottomrule[1pt]
\end{tabular}
\begin{tablenotes}
\footnotesize
\item[1] 90 means the $90^{th}$ percentile of the data.
\item[2] Policy without encoder.
\item[3] Policy with encoder.
\end{tablenotes}
\end{threeparttable}
\vspace{-0.4cm} \end{table}

% table4
\begin{table}[ht]
\centering
\caption{FAP VS. AP}
\label{adpfvsnf}
\begin{threeparttable}

\resizebox{\linewidth}{!}{
\begin{tabular}{lcccccc} 
\toprule[1pt]

       & \multicolumn{3}{c}{\textbf{Force}(N)}                    & \multicolumn{3}{c}{\textbf{Torque}(N·m)}                     \\ 
       & \textbf{Mean}          & \textbf{Maximum}       & \textbf{90}\tnote{1} & \textbf{Mean}          & \textbf{Maximum}       & \textbf{90}\tnote{1}  \\ 
\midrule
\textbf{AP} & 20.2          & 51.9          & 37.2          & 3.77          & 10.5          & 7.01           \\ 
\textbf{FAP}  & \textbf{18.7} & \textbf{44.0} & \textbf{37.0} & \textbf{3.36} & \textbf{8.29} & \textbf{6.56}  \\

\bottomrule[1pt]
\end{tabular}}
\begin{tablenotes}
\footnotesize
\item[1] 90 means the $90^{th}$ percentile of the data
\end{tablenotes}
\end{threeparttable}
\vspace{-0.4cm} \end{table}

% table5
\begin{table}[ht]
\centering
\caption{TDO VS. ours}
\label{trovsours}
\begin{threeparttable}
\resizebox{\linewidth}{!}{
\begin{tabular}{cccccccc} 
\toprule[1pt]
&& \multicolumn{2}{c}{\textbf{Cabinet A}} & \multicolumn{2}{c}{\textbf{Cabinet B}} & \multicolumn{2}{c}{\textbf{Cabinet C}}\\ \midrule
 
\multicolumn{2}{c}{\textbf{Method}} & \textbf{TDO} & \textbf{ours} & \textbf{TDO} & \textbf{ours} & \textbf{TDO} & \textbf{ours}\\ \midrule
 
% \multicolumn{2}{c}{\textbf{Success rate}} & \textbf{100\%(30/30)} & \textbf{100\%(30/30)} & 76.7\%(23/30) & \textbf{100\%(30/30)}&
% 83.3\%(25/30) & \textbf{100\%(30/30)}\\ \midrule
  \multicolumn{2}{c}{\textbf{Success rate}} & \textbf{100\%} & \textbf{100\%} & 76.7\% & \textbf{100\%}&
83.3\% & \textbf{100\%}\\ \midrule

\multirow{3}{*}{\shortstack{\textbf{Force}\\(N)} } &\textbf{50}\tnote{1} & \textbf{11.5} & 13.4 & 18.4 & \textbf{14.8} & 19.5 &  \textbf{18.7} \\ & \textbf{90}\tnote{2} & \textbf{25.4} & 38.9 & 47.2 & \textbf{34.5} & \textbf{38.4} &  39.0 \\  & \textbf{Max}\tnote{3}& \textbf{55.4} & 54.1 & 94.9 & \textbf{62.8} & 73.1 &  \textbf{54.1} \\ \midrule

\multirow{3}{*}{\shortstack{\textbf{Torque}\\(N·m)}} &\textbf{50}\tnote{1} & \textbf{1.50} & 2.42 & 3.01 & \textbf{1.56} & 2.52 &  \textbf{2.34} \\ & \textbf{90}\tnote{2} & \textbf{3.67} & 7.52 & 8.76 & \textbf{4.34} & 6.05 &  \textbf{5.10} \\  & \textbf{Max}\tnote{3}& \textbf{9.00} & 11.4 & 20.2 & \textbf{8.02} & 15.9 &  \textbf{7.73} \\ 
\bottomrule[1pt]
\end{tabular}
}

\begin{tablenotes}
\footnotesize
\item[1] 50 means the $50^{th}$ percentile of the data
\item[2] 90 means the $90^{th}$ percentile of the data
\item[3] Max means the maximum of the data
\end{tablenotes}
\end{threeparttable}
\vspace{-0.4cm} \end{table}

\begin{figure}[t]
    \centering
    \includegraphics[width=4cm]{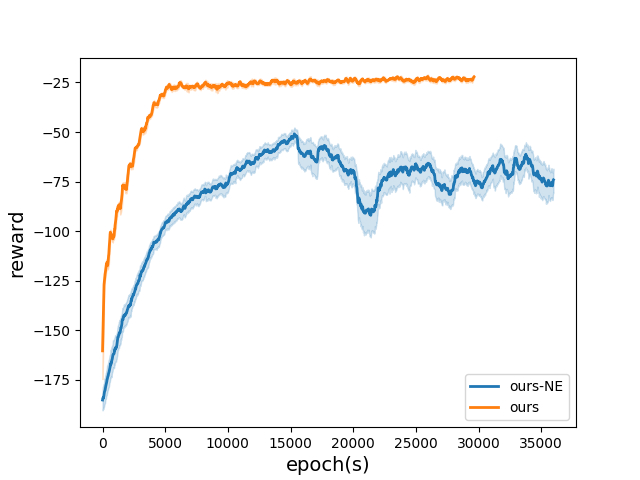}
    \label{fig:traincurve_withencoder}
    \caption{ This figure shows the policy with encoder not only takes less time to convergence but also obtains  the higher reward compared to the policy without encoder.}
\vspace{-0.4cm}
\end{figure}

\begin{figure}[t]
    \centering
    \subfigure[The estimation error of r and $\theta$]{
    \includegraphics[width=4cm]{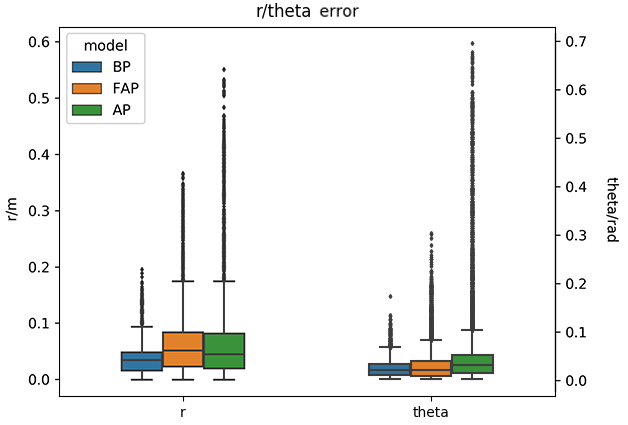}
    \label{fig:fap_rtheta}
    }
    \subfigure[The force-torque data]{
    \includegraphics[width=4cm]{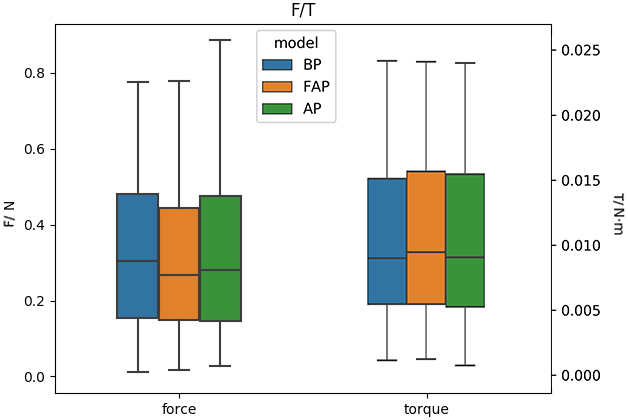}
    \label{fig:fap_force}
    }
    \caption{The boxplots of the results among BP, FAP and AP. Fig(a) shows the there is a gap between the performance of AP with BP, and FAP eliminates the outliners. Fig(b) shows that FAP a little bit improves the force performance of AP.}
    \label{fig:fap}
\vspace{-0.4cm}
\end{figure}

\section{EXPERIMENTAL RESULTS}\label{section:results}

\subsection{Ablation Study}\label{ablation}
We conduct ablation studies by removing individual components of our framework. We evaluate the methods by the force-torque magnitude during the door opening, with minor force-torque magnitude related to a better method. Furthermore, we count a success trial when the cabinet door is opened at an angle greater than 45 degrees. In the simulation, we take the average performance on 20 trails with the same seed for each experiment. In the real-world environment, for each experiment, we test on the same cabinet door with a random initial grasping position ($\theta$ in -15° to 15°) for ten times. Studies (1)(2)(3)(4) are conducted in simulation, while Study (5) is conducted in both simulation and real world.

\subsubsection{Training on Single Door VS. Training by Domain Randomization}
To verify the necessity of random sampling in the environment parameters domain, we train a policy based on a single door in which parameters are average of the domain. The results shown in Table. \ref{singlevsdomain} reflect that the policy trained by domain randomization leads to a better performance by a wide margin.

\subsubsection{Policy With Encoder VS. Policy Without Encoder}
We form an new netowrk $\varrho$ to directly inform the feature from the action-state sequence without the supervision of $\boldsymbol{z}$. The structure of $\varrho$ is the same with $\rho$. We simultaneously train the $\varrho$ and a new policy $\pi^{*}$ (\ref{eepolicy}), which is an end-to-end policy without environment parameters and encoder. We compare the performance of $\pi^{*}$ with the base policy, in order to validate the effectiveness of the encoder $\delta$
\begin{equation}
\boldsymbol{a_t} = \pi^{*}(\varrho(\boldsymbol{x_{t-n:t}}))
\label{eepolicy}
\end{equation}

As the Fig. \ref{fig:traincurve_withencoder} shows, the policy with encoder (our base policy) not only takes less time to convergence but also obtains the higher reward compared to the policy without encoder. The results in Table. \ref{encoderwithout} show that the estimations from the policy with encoder is more accurate. 

\subsubsection{FAP VS. AP}
In the simulation, we take the performance of BP as the reference, and then we test the performance of FAP and AP. The boxplots in Fig. \ref{fig:fap} demonstrate that there is a gap between the performance of AP with BP, for the impact of noise we discussed in Section \ref{subsectionfine}, and FAP eliminates the outliners of $\hat{r}$ and $\hat{\theta}$. Additionally, we compare the performance of FAP and AP in the real world and the results in Table. \ref{adpfvsnf} prove the effectiveness of FAP.

\subsection{Comparison with Traditional Method}
In this section, we prepare three cabinet doors (Fig. \ref{teaser}) with different properties to examine the generalization of our method (FAP). We also implement the traditional door opening method (TDO) \cite{karayiannidis2016adaptive} for baseline comparison. Both methods are tested 30 times on each cabinet door with a random initial grasping position ($\theta$ in -15° to 15°). As it is a hand-engineering method, we initiate the TDO to ensure a robust performance on cabinet door A, and then we maintain the parameters for the experiments of cabinet doors B and C. 

Table. \ref{trovsours} shows the success rate and recorded force-torque data. We see both our method and TDO achieve the 100\% success rate on cabinet door A. TDO performs better than our method for the sake of proper initial parameters. However, things turn around when we go on with cabinet doors B and C. TDO reveals the weak performance and even encounters failed trials, while our method maintains the 100\% success rate and produces a minor force-torque magnitude. 

In this way, our method has the capability of operating different doors in the real world. Furthermore, compared to the hand-engineering method, our method shows a more robust performance without any manual parameter adjustment.

% \subsection{Case in Daily Life}
% ?

% \begin{figure}[t]
% \centering
% \includegraphics[scale=0.5]{2vs6_new.png}
% \caption{The curve of the training process}
% \label{2vs6fig}
% \end{figure}

% \begin{figure}[t]
% \centering
% \includegraphics[scale=0.5]{ours_NE.png}
% \caption{The curve of the training process}
% \label{encoderwithoutfig}
% \end{figure}

% \begin{figure}[t]
% \centering
% \includegraphics[scale=0.35]{based_ft.png}
% \caption{The curve of the training process}
% \label{baseftfig}
% \end{figure}

% \begin{figure}[t]
% \centering
% \includegraphics[scale=0.35]{ppa_rtheta.png}
% \caption{The curve of the training process}
% \label{adpnfrthetafig}
% \end{figure}

\section{CONCLUSIONS}

In this paper, we introduce a learning-based framework for robot adaptive manipulation of the object with a revolute joint in unstructured environments. We train the policy in the simulation with the encoded environment parameters, and a fine-tuned adaptation module helps the policy development in the real world. We design a series of experiments, and the results demonstrate the efficacy and efficiency of our framework.
In future work, we will focus on identifying the target object and attaching the effector to it to fundamentally guide our framework to be applied in the real world.

\clearpage
\newpage

\section{Supplementary Material}

\subsection{Enviromment Parameters and Encoder Training}
\label{section:encodertraining_2}
To depict the various configurations of the cabinet doors, we specify the environment parameters $\boldsymbol{e}$ subject to daily life, with their value range shown in Table. \ref{environmentparameters}.

\begin{table}[htbp]
\centering
\caption{Details of environment parameters}
\label{environmentparameters}
\begin{tabular}{lll}
\toprule[1pt]
            & \textbf{Value Range}            & \textbf{Units}     \\
\midrule
length      & (0.28, 0.32)           & $m$          \\
width       & (0.2,     0.85)        & $m$          \\
height      & (0.2,     0.4)         & $m$          \\
thicc       & (0.01, 0.03)           & $m$          \\
density     & (300, 3000)            & $kg/m^3$     \\
damping     & (0.01, 0,08)           & $N·s/m$      \\
friction    & (0.001, 0.02)          & /          \\
position\_x & (0.45, 0.55)           & $m$          \\
position\_y & (-0.05, 0.05)          & $m$          \\
position\_z & (-0.05, 0.05)          & $m$          \\
quaternion\_w & (-0.1284, 1)          & /          \\
quaternion\_x & (-0.0489, 0.0489)     & /          \\
quaternion\_y & (-0.0489, 0.0489)     & /          \\
quaternion\_z & (0.989, 1)            & /          \\
$\theta$      & (-0.3, 0.3)            & $rad$        \\
liner velocity  & (-0.3, -0.05)      & $m/s^2$      \\
\bottomrule[1pt]
\end{tabular}
\vspace{-0.4cm} \end{table}

% table3
\begin{table}[t]
\centering
\caption{Results of 2-Dof Policy VS. 6-Dof Policy}
\label{2vs6}
\begin{threeparttable}
\begin{tabular}{cccc} 
\toprule[1pt]
                                                &           & \textbf{6-DoF} & \textbf{2-DoF}\\ \midrule
    \textbf{Success rate}                       &       & 15\%(3/20) & \textbf{100\%(20/20)} \\ \midrule
    \multirow{2}{*}{\shortstack{\textbf{Error of}\\  \textbf{velocity}(m/s)}} & \textbf{Mean}        & 47.9 & \textbf{6.23}\\
                                                & \textbf{90}\tnote{1} & 69.6 & \textbf{16.4}\\ \midrule
    \multirow{3}{*}{\shortstack{\textbf{Force} \\ (N*$e^{-2})$}}   & \textbf{Mean}        & 11.3 & \textbf{5.97}\\
                                                & \textbf{Maximum}     & 73.5 & \textbf{59.6}\\
                                                & \textbf{90}\tnote{1} & 17.7 & \textbf{14.8}\\ \midrule
    \multirow{3}{*}{\shortstack{\textbf{Torque}\\(N·m*$e^{-2}$)}}   & \textbf{Mean}     & 5.62 & \textbf{3.32}\\
                                                & \textbf{Maximum}     & 28.23& \textbf{20.9}\\
                                                & \textbf{90}\tnote{1} & 7.60 & \textbf{7.46}\\

\bottomrule[1pt]
\end{tabular}
\begin{tablenotes}
\footnotesize
\item[1] 90 means the $90^{th}$ percentile of the data
\end{tablenotes}
\end{threeparttable}
\vspace{-0.2cm} \end{table}

% table2
\begin{table}[t]
\centering
\caption{Results of $r \& \theta$ Supervised VS. Velocity Supervised}
\label{rthetavsvelocity}
\setlength{\tabcolsep}{10pt}
\begin{threeparttable}
\begin{tabular}{lcccc} 
\toprule[1pt]

& \multicolumn{4}{c}{\textbf{Error}(*e$^{-2}$)}                                                      \\

& \multicolumn{2}{c}{\textbf{r}(m)}                             & \multicolumn{2}{c}{$\boldsymbol{\theta}$(rad)}          \\

& \textbf{Mean}          & \textbf{90}\tnote{1} & \textbf{Mean}          & \textbf{90}\tnote{1}  \\
\midrule
$\boldsymbol{r \& \theta}$          & 10.5          & 16.7                              & 12.0          & 30.2                               \\
\textbf{Velocity}               & \textbf{4.56} & \textbf{12.7}                     & \textbf{2.17} & \textbf{4.41}                      \\

\bottomrule[1pt]
\end{tabular}
\begin{tablenotes}
\footnotesize
\item[1] 90 means the $90^{th}$ percentile of the data.

\end{tablenotes}
\end{threeparttable}
\vspace{-0.4cm} \end{table}

\subsection{Dimension Reduction of Searching Space}\label{subseactionsearching_2} 
% Rather than free-joint objects, objects with revolute joints have kinematical structures that cannot be violated, which means the effector randomly interacting would cause damage to either the effector or the objects. Therefore, searching for reinforcement learning in the whole action space is time-consuming and nonsense. Firstly, we contract the searching space by physically modeling the manipulation task.

In the cabinet door opening case, according to Fig. \ref{space}, we make the following assumptions: (1) The effector is a robotic gripper and has grasped at the cabinet door. The YZ-plane of the gripper is parallel to the YZ-plane of the cabinet door and between z-axis of the gripper $\{z_g\}$ and y-axis of the cabinet door $\{y_d\}$ there is an angle $\theta$. (2) The distance from the gripper’s end center $O_g$ to the z-axis of the cabinet door $\{z_d\}$ is $r$. (3) In the ideal door opening process, the cabinet door rotates uniformly with the angular velocity 
\begin{equation}
    \boldsymbol{\omega}_d = \begin{bmatrix}
  0& 0 & \omega
\end{bmatrix}^T
\label{omega_d}
\end{equation}
while the gripper is stationary with respect to the cabinet door during the process.

We set the desired gripper’s velocity $\widetilde{\mathbf{v}}$ with respect to the gripper frame $\{g\}$ as 
\begin{equation}
    \widetilde{\mathbf{v}} = \begin{bmatrix}
  \mathbf{v}_g &  \boldsymbol{\omega}_g
\end{bmatrix}^T
\label{vhatd}
\end{equation}
including the linear velocity
\begin{equation}
\mathbf{v}_g = \begin{bmatrix}
v_{gx} & v_{gy} & v_{gz}
\end{bmatrix}^T
\label{v_g}
\end{equation}
and the angular velocity 
\begin{equation}
    \boldsymbol{\omega_g = }
    \begin{bmatrix}
\omega_{gx} & \omega_{gy} & \omega_{gz}
\end{bmatrix}^T
\label{omega_g}
\end{equation}

According assumption (1) and (2), we can figure out the gripper’s rotation matrix $\mathbf{R_{dg}}$ with respect to the cabinet door frame $\{d\}$
\begin{equation}
\mathbf{R_{dg}}=\begin{bmatrix}
 -1 & 0& 0\\
 0 & -sin\theta & -cos\theta\\
  0&-cos\theta  &sin\theta
\end{bmatrix}
\label{r_dg}
\end{equation}
and we set the gripper’s translation matrix $\mathbf{T_{dg}}$ with respect to frame $\{d\}$ 
\begin{equation}
    \mathbf{T_{dg}} =\begin{bmatrix}
0 & r  &z_{dg}
\end{bmatrix}^T
\label{t_dg}
\end{equation}

According to the assumption (3), we find 
\begin{equation}
    \boldsymbol{\omega_{dg}} = \boldsymbol{\omega_d} =     \begin{bmatrix}
  0& 0 & \omega
\end{bmatrix}^T
\label{omega_dg}
\end{equation}
where $\boldsymbol{\omega_{dg}}$ is the desired angular velocity of the gripper with respect to the frame $\{d\}$, and then the desired liner velocity of the gripper $\mathbf{v_{dg}}$ with respect to the frame $\{d\}$  can be calculated as
\begin{equation}
   \mathbf{v_{dg}} = \boldsymbol{\omega_{dg}} \times r = \boldsymbol{\omega_{dg}} \times \mathbf{T_{dg}} = \begin{bmatrix}
-\omega r & 0 & 0
\end{bmatrix}^T
\label{v_dg}
\end{equation}

From Eq. (\ref{r_dg})-(\ref{v_dg}) , we can get 
\begin{equation}
    \mathbf{v_g} = \mathbf{R_{dg}^T} \mathbf{v_{dg}} = \begin{bmatrix}
\omega r & 0 & 0
\end{bmatrix}^T
\label{v_gr}
\end{equation}
\begin{equation}
    \boldsymbol{\omega_g} = \mathbf{R_{dg}^T} \boldsymbol{\omega_{dg}} = \begin{bmatrix}
0 & -\omega cos\theta& \omega sin\theta
\end{bmatrix}^T
\label{omega_gtheta}
\end{equation}

Then with Eq. (\ref{vhatd})(\ref{v_g})(\ref{omega_g})(\ref{v_dg})(\ref{omega_gtheta}), we conduct a new expression of $\mathbf{\widetilde{v}}$ represented by $r, \theta$
\begin{equation}
    \mathbf{\widetilde{v}} =\omega \begin{bmatrix}
    r & 0 & 0 & 0 & -cos\theta &sin\theta
    \end{bmatrix}^T
\label{v_drtheta}
\end{equation}

Based on Eq.(\ref{v_drtheta}), we convert the searching space of DRL from $\mathbf{\widetilde{v}} \in \mathbb{R}^{6\times1}$ to $\begin{bmatrix}
r & \theta
\end{bmatrix}^T \in \mathbb{R}^{2\times1}$, which speeds up the model’s training process and raises up the model’s performance.

\subsection{Ablation Study}\label{ablation_2}

\begin{figure}[t]
\centering
\includegraphics[width=0.6\linewidth]{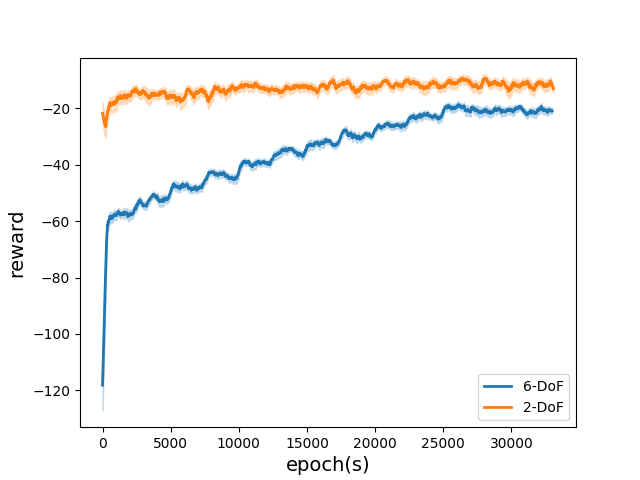}
\caption{2-DoF policy collects higher reward at 3k epochs than 6-Dof when both models come to convergence.}
\label{fig:traincurve_2vs6}
\vspace{-0.7cm}
\end{figure}

\noindent
\textbf{2-DoF Policy VS. 6-DoF Policy}
We compare the policy trained in the dimension-reduced searching space (2-DoF policy) with in the whole searching space (6-DoF policy). The searching space of 6-DoF policy is the velocity of the gripper $\mathbf{v_g}$ so that in this experiment the reward function at time $t$ is redefined as 
\begin{equation}
\begin{aligned}
      R(\boldsymbol{x_t}, \boldsymbol{a_{t-1}}) =  - & (k_1||\mathbf{v_g}-\mathbf{\widetilde{v}_g}||^2 + k_2|F_y| +    k_3|F_z| + \\ &  k_4|\tau_x| + k_5|\frac{\tau_z}{\tau_y}+tan\theta|) 
\end{aligned}
\label{reward_rt}
\end{equation}
where $\mathbf{\widetilde{v}_g}$ is the ideal velocity of the gripper during the door opening process.  

As shown in Fig.\ref{fig:traincurve_2vs6}, the 2-DoF policy collects higher reward at 3k epochs when both models come to convergence. Table. \ref{2vs6} shows that the 2-DoF policy performs better in both force-torque data and success rate. The low success rate of the 6-DoF indicates that training without dimension-reduced searching space is impracticable even the model has been convergent.

\noindent
\textbf{$\bf{r \& \theta}$ Supervised VS. Velocity Supervised}
We experiment to compare the original reward function and the redefined reward function in the ablation study (1). We find that the policy with $r \& \theta$ supervised produces more accurate $\hat{r}$ and $\hat{\theta}$ according to the Table. \ref{rthetavsvelocity}, because this manner logically reflects the analysis in Section \ref{subseactionsearching}.

\end{document}